\documentclass[letterpaper, 10 pt, conference]{ieeeconf}

\IEEEoverridecommandlockouts                              
\usepackage{graphicx}
\usepackage{subcaption}
\usepackage{multirow}
\usepackage{url}
\usepackage[section]{placeins}

\usepackage{amsmath} 
\usepackage{amssymb}
\usepackage{hyperref}
\usepackage{xurl}

\title{\LARGE \bf
A Conversational Framework for Human-Robot Collaborative Manipulation with Distributed Generative AI models
}
\author{Arash Ghasemzadeh Kakroudi$^{1}$ and Roel Pieters$^{2}$%
\thanks{$^{1}$Automation Technology and Mechanical Engineering, Tampere University, 33720, Tampere, Finland, {\tt\small arash.ghasemzadehkakroudi@tuni.fi}}%
\thanks{$^{2}$Automation Technology and Mechanical Engineering, Tampere University, 33720, Tampere, Finland, {\tt\small roel.pieters@tuni.fi}}%
}

\begin{document}

\maketitle

\begin{abstract}
This paper presents a distributed conversational framework for human-robot collaborative manipulation that integrates local language and vision-language models (VLMs) with a Robot Operating System 2 (ROS~2)-based execution stack. Language understanding, visual grounding, orchestration, and motion execution run as separate ROS~2 nodes, enabling flexible deployment across distributed hardware while maintaining a responsive control loop. From free-form user commands, the system generates structured action requests for pick, place, and handover. It uses a VLM to return image-space targets, which are converted into metric robot-frame goals using depth and calibration. A web dashboard exposes intermediate intent and grounding overlays (pixel, depth, and robot-frame) and requires explicit operator confirmation before any motion is executed. Experiments on a Franka FR3 platform evaluate end-to-end task reliability and latency under increasing working table scene ambiguity and compare alternative LLM/VLM configurations in the same pipeline. Code and full documentation are available at \href{https://github.com/cogrob-tuni/franka-llm}{github.com/cogrob-tuni/franka-llm}.
\end{abstract}

\section{INTRODUCTION}
Robotic manipulation has made strong progress in planning, control, and perception, but natural-language interaction in real work settings is still challenging~\cite{Shen2026TiPToP}. User commands can be incomplete, scene descriptions can be ambiguous, and requested objects may look similar from the robot's viewpoint~\cite{Gkournelos2024LLMAssembly,Liu2026Copilot}. At the same time, recent language and vision models make interaction easier by allowing people to communicate with robots in everyday language, which is valuable for practical human–robot collaboration~\cite{Liu2024VisionAI,Chen2026SteerableVLA}.

\begin{figure}
    \centering
    \includegraphics[width=1\linewidth]{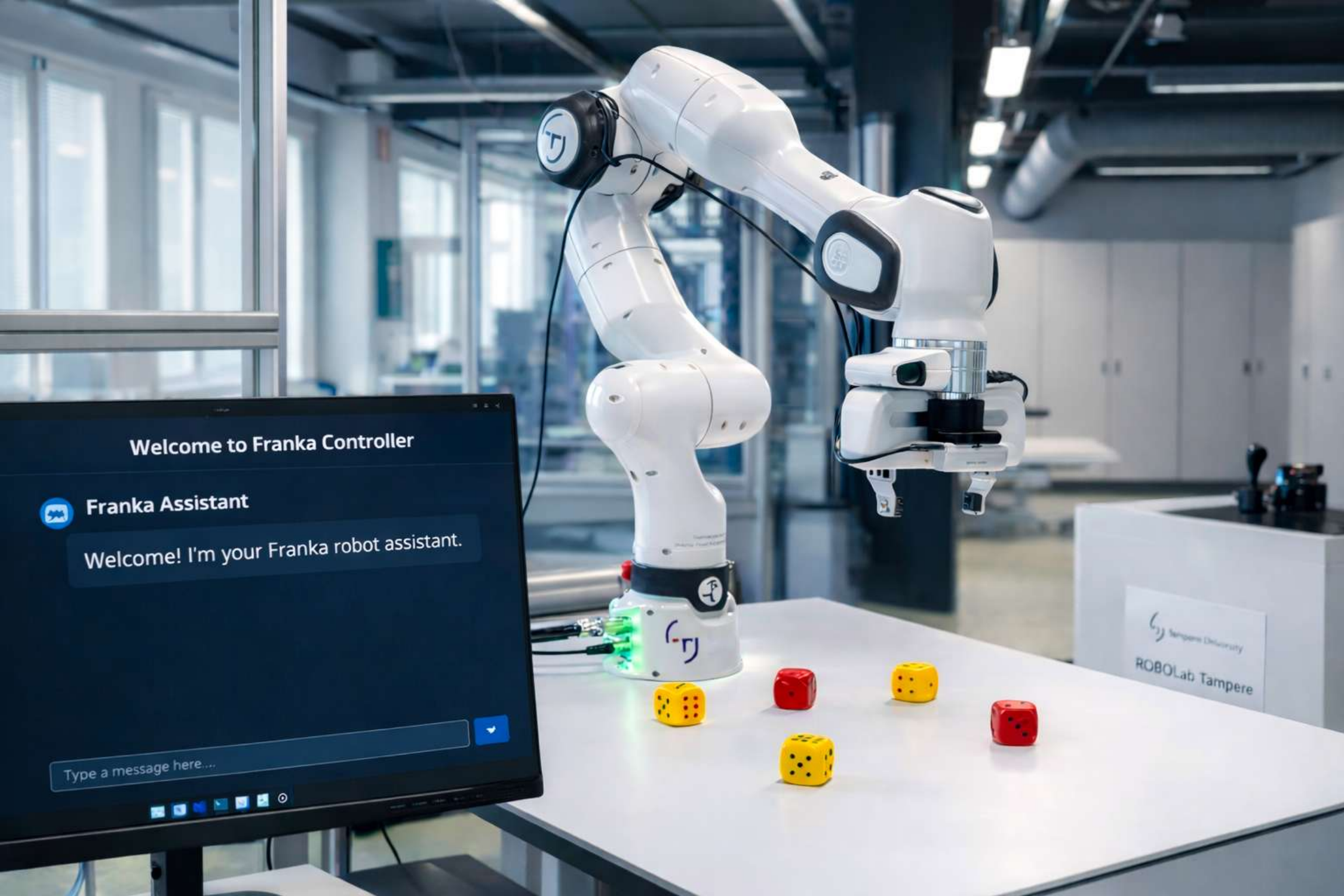}
    \caption{System overview and experimental setup.}
    \label{fig:intro_img}
    \vspace{-1em}
\end{figure}

In this paper, we present a distributed conversational framework for human–robot collaborative manipulation. The system is implemented as a ROS~2 multi-agent pipeline where language understanding, visual grounding, orchestration, and motion execution run as separate nodes, each of which can be deployed on a separate device in a distributed setup~\cite{Macenski2022,Macenski2023Composition}. A web-based dashboard accepts free-form user commands, displays interpreted action intents and grounding overlays (pixel/depth/robot-frame), and provides an explicit confirmation dialog before any motion. In our default deployment, the large language model (LLM)-based planner runs on an edge device while the VLM-based visual grounding runs on a workstation graphics processing unit (GPU), but the architecture supports single-machine and fully distributed placements depending on available hardware~\cite{Wang2025}. Fig.~\ref{fig:intro_img} shows the overall system and experimental setup (Franka FR3, dashboard, and worktable).

The key risk is the reliability of model outputs. LLMs are known to hallucinate, meaning they may generate plausible but incorrect content~\cite{Ji2022HallucinationSurvey,Huang2023LLMHallucination}. In robotics, this is not only a quality issue but also a safety issue: a wrong interpretation can lead to an unintended physical action. For this reason, our framework does not allow free-form model output to directly trigger robot motion. Instead, language and vision outputs are translated into structured action requests, and every executable action is presented to the user for confirmation before execution. This human confirmation step is the main safety barrier in the interaction loop and follows recent work on proactive safety prioritization for close human-robot interaction (HRI)~\cite{Maithani2025CBF}.
To make this practical, we use a distributed multi-agent architecture where language understanding, visual grounding, orchestration, and motion execution are separated into ROS~2 nodes that can be deployed across distributed hardware~\cite{Macenski2022,Macenski2023Composition}. 

We position the system relative to two recent trends: LLM-centered instruction understanding for HRI tasks~\cite{Gkournelos2024LLMAssembly,Liu2026Copilot} and VLM-driven open-vocabulary grounding for manipulation~\cite{Chen2026SteerableVLA,Zhou2026MARVL}. Our contribution is not a monolithic end-to-end policy; instead, we emphasize an inspectable, distributed ROS~2 pipeline with explicit confirmation and flexible deployment across heterogeneous hardware, including edge devices and workstation GPUs~\cite{Wang2025}. 
In this work, we present the architecture for pick, place, and handover tasks and evaluate whether it can remain accurate, responsive, and interpretable under increasing scene ambiguity.

\noindent\textbf{Contributions:}
\begin{itemize}
    \item A distributed ROS~2 multi-agent pipeline that separates language, visual grounding, orchestration, and motion execution, enabling flexible single-machine or multi-device deployments and easy adaptation to available hardware.
    \item A web dashboard that exposes intermediate intent and grounding overlays and enforces an explicit operator confirmation gate before any motion.
    \item An open-vocabulary grounding-to-robot-frame target generation pipeline (pixel/depth/robot-frame) for pick, place, and handover primitives.
    \item An evaluation under increasing scene ambiguity, comparing alternative language/vision model pairings within the same modular architecture.
\end{itemize}

\section{Related work}
Recent work has shown that LLMs can help translate free-form user instructions into structured robot goals, but also surfaces persistent interaction challenges such as underspecification and ambiguity in collaborative settings. In assembly-oriented HRC, Gkournelos et al.~\cite{Gkournelos2024LLMAssembly} propose an LLM-based approach to support seamless collaboration, while Liu et al.~\cite{Liu2026Copilot} introduce a Copilot framework that integrates LLM reasoning with BMI signals to enhance human--robot interaction. Vision-centric system work also highlights the role of rich perception for collaborative assembly, e.g., Vision AI-based human--robot assembly driven by autonomous robots~\cite{Liu2024VisionAI}.

System-oriented research increasingly emphasizes end-to-end pipelines that connect instruction understanding to task-and-motion execution with clear modular boundaries. Shen et al.~\cite{Shen2026TiPToP} present TiPToP, which integrates open-vocabulary perception signals with a planner and highlights failure analysis across components. Related instruction-to-action pipelines also commonly decompose commands into ordered atomic actions, which improves traceability and simplifies debugging when execution diverges from intent~\cite{Sharma2026Instruct2Act}.

Complementary to these pipelines, hierarchical VLM-to-policy stacks separate high-level reasoning from low-level control, but their performance hinges on how expressively the high-level model can steer the low-level policy. Chen et al.~\cite{Chen2026SteerableVLA} identify the natural-language-only interface as a bottleneck and propose Steerable Policies: vision--language--action (VLA) policies trained on synthetically generated steering commands spanning multiple abstractions, including task-level language, subtask instructions (e.g., ``reach''), atomic motions (e.g., ``move left''/``open gripper''), and grounded 2D conditioning such as points and gripper traces, as well as their combinations. They further show two hierarchical variants where either a learned embodied reasoner outputs reasoning followed by steering commands, or an off-the-shelf VLM uses in-context learning to choose the appropriate abstraction online to resolve ambiguity in cluttered scenes~\cite{Chen2026SteerableVLA}. In our system, the coordinator targets the same interface problem at the system-integration layer: rather than emitting a single free-form instruction to a monolithic policy, it splits each user utterance into typed sub-requests (dialogue/clarification, open-vocabulary grounding, and motion execution) with explicit inputs/outputs. This ``request routing'' makes intermediate intent and grounding inspectable and supports operator oversight by separating ``what'' (language) from ``where'' (vision) and ``how'' (robot control), aligning with modular design principles emphasized in TiPToP-style analysis~\cite{Shen2026TiPToP}.

On the perception side, object grounding has shifted from closed-vocabulary detectors (YOLO family) toward large VLMs that support open-vocabulary, language-conditioned localization and reasoning. Steerable VLA policies and MARVL demonstrate compositional referring-expression grounding, multi-stage spatial guidance, and active perception strategies to resolve ambiguity~\cite{Chen2026SteerableVLA,Zhou2026MARVL}. In contrast, YOLO-style detectors provide fast and reliable inference for a fixed label set but require additional labeled data or retraining to expand vocabulary~\cite{Redmon2016YOLO}.

Practical deployment constraints remain central for interactive HRI systems that integrate LLMs/VLMs. The on-device AI survey by Wang et al.~\cite{Wang2025} emphasizes memory footprint, latency, and privacy considerations that affect whether models should run on edge devices, workstations, or remote servers. In parallel, safety-focused work such as hierarchical control-barrier methods for close HRI~\cite{Maithani2025CBF} and surveys on hallucination behavior~\cite{Ji2022HallucinationSurvey,Huang2023LLMHallucination} motivate explicit oversight mechanisms when model outputs can trigger physical motion.

Compared to recent assembly-centric LLM frameworks~\cite{Gkournelos2024LLMAssembly,Liu2026Copilot} and end-to-end VLM/VLA pipelines~\cite{Chen2026SteerableVLA,Zhou2026MARVL}, our focus is system integration: a distributed ROS~2 architecture with typed intermediate representations (intent, grounding, and execution requests) that remain inspectable throughout the pipeline. This separation makes it easier to debug failures, swap models, and adapt deployment (single machine vs.~fully distributed) without changing downstream control.
\section{Methodology}
\begin{figure*}[htp]
\centering
{
  \includegraphics[width=0.92\textwidth]{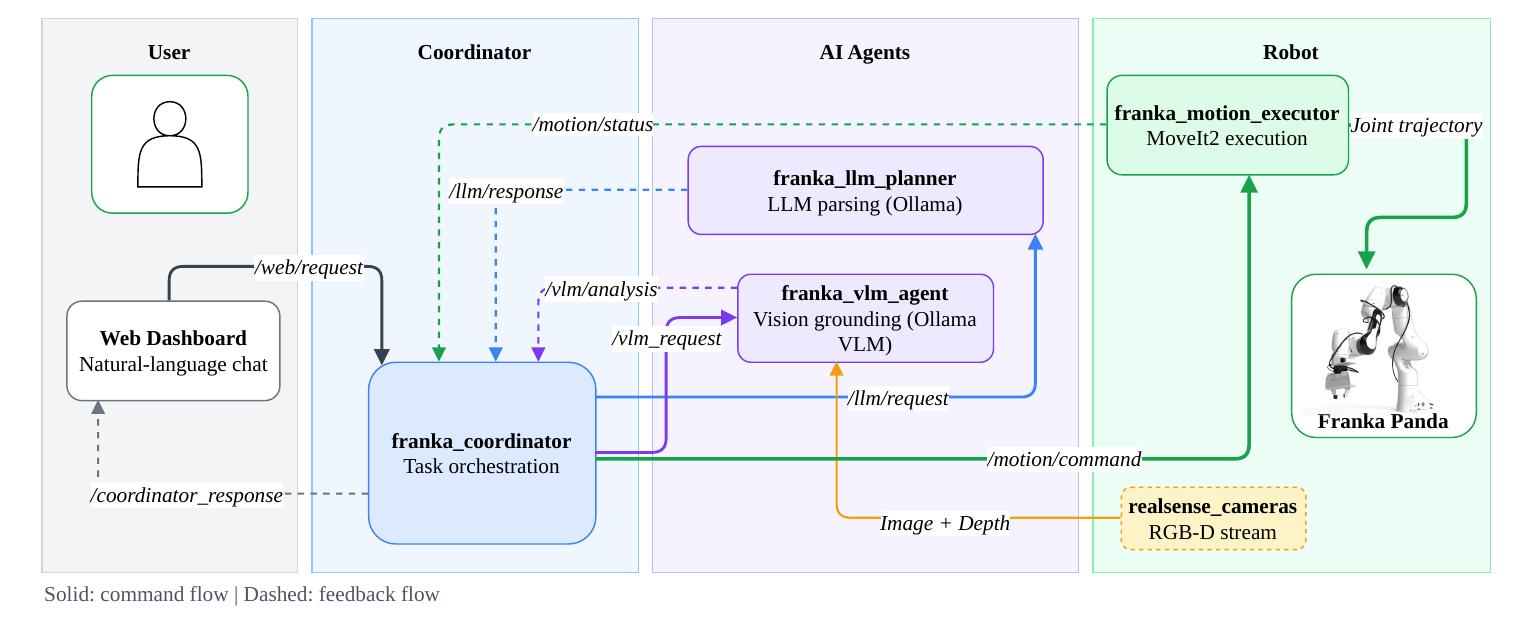}
}\hfill
\caption{High-level data flow architecture of the proposed distributed framework\label{fig:Block_diag}
}
\end{figure*}

\subsection{Distributed Hardware Architecture and Multi-Agent Coordination}
The proposed system is designed as a distributed robotic manipulation framework in which perception, language understanding, coordination, and motion execution are decoupled into separate ROS~2 nodes~\cite{Macenski2022,Macenski2023Composition}. This design allows the framework to be deployed either on a single node or across multiple computing devices, depending on available hardware resources. In our implementation, the Franka Emika FR3 robot~\cite{Franka2026} is combined with an RGB-D (RGB + depth) end-effector camera, a web-based user interface, a coordination layer, and two AI modules for language and vision processing.

Fig.~\ref{fig:Block_diag} illustrates the high-level data flow. The user interacts with the system through a web dashboard using natural language commands. These commands are received by the central coordinator, which routes requests to the language and vision agents and finally forwards validated motion commands to the robot controller. The architecture separates high-level reasoning from low-level motion execution, which improves modularity, simplifies debugging, and makes it possible to move computationally expensive components to external hardware when required.

The framework follows a multi-agent design in which each ROS~2 node has a clearly defined role. A language-understanding node interprets natural language commands and converts them into structured action requests. A vision-grounding node performs scene understanding and object grounding from RGB images. The central node directs agent requests, combines visual data with depth and calibration, and outputs robot-frame targets for execution. Finally, a motion-execution node generates and executes robot trajectories through MoveIt2~\cite{MoveIt2026}. The coordinator is the key integration component: it receives user requests from the web handler, dispatches language and vision queries, combines results into executable targets, and manages the safety-oriented confirmation loop. For actions such as pick, place, handover, and go-home, the system first presents the interpreted action and target information to the user through the dashboard. Motion is executed only after explicit user approval.

\subsection{Local Model Deployment and Privacy}

Both generative AI models are executed locally through Ollama~\cite{Ollama2026}. In the default deployed setup, the language understanding module uses \mbox{ministral-3:8b} on the NVIDIA Jetson AGX Orin 64 GB platform~\cite{Mistral8B2026}, while the visual grounding module uses Qwen2.5-VL~32B on the RTX~6000 workstation~\cite{Qwen25VL2026}. Running the models locally removes the need to send user commands, camera images, or robot-related data to external cloud services. This is particularly important in human-robot collaboration settings, where visual observations may contain sensitive workspace information and where predictable system latency is required for interactive operation.

Local deployment also gives flexibility in hardware placement. The full system can be executed on one machine for compact setups, while the LLM and VLM nodes can also be distributed to more capable devices, such as a workstation GPU server. This is useful because the VLM is computationally more demanding than the language planner, whereas the motion execution and coordination nodes benefit from staying close to the robot control stack. In addition, the system behavior is controlled through a shared \texttt{config.yaml} file. This file defines the active LLM and VLM models, Ollama endpoints, camera topics, calibration offsets, robot motion constants, placement distances, and debug settings. As a result, the same software pipeline can be reconfigured without changing the source code.

\subsection{Natural Language Processing Pipeline}

Natural language understanding is performed by the \mbox{ministral-3:8b} model through the LLM coordinator node~\cite{Mistral8B2026}. The purpose of this module is to map users' input into a compact and structured decision format. Each incoming command is classified into an action type such as \emph{pick}, \emph{place}, \emph{handover}, or \emph{go\_home}, together with the relevant parameters such as object identity, target location, or placement intent.
This structured representation is important for reliable downstream execution. Rather than allowing the LLM to directly control robot motion, the planner only selects which subsystem should be activated and what information should be extracted. Greetings or simple conversational requests can be answered directly, whereas object-centered commands are routed to the vision module for grounding. For place requests, the parsed intent can also include directional relations such as left, right, top, or bottom with respect to a reference object. In such cases, the coordinator applies a configurable offset distance of 8~cm to generate the final placement target.

\subsection{Vision-Language Grounding}

Visual grounding is handled by the Qwen2.5-VL~32B model operating on RGB input from the end-effector Intel RealSense camera~\cite{IntelRealSense2026}. For location-grounding requests, which are required for place tasks, the VLM is prompted to return a direct center pixel in the text format \texttt{FOUND at pixel (u, v)}. In our implementation, the coordinator parses the free-text output using a regex cascade that supports both pixel coordinates and normalized values ($u, v \leq 1.0$). When normalized values are returned, they are rescaled to the current image resolution to ensure consistent target localization from the VLM. OpenCV~\cite{OpenCV2000} is used for image handling and overlay generation in the debugging and operator-feedback pipeline.

The extracted center pixel $(u_c, v_c)$ is fused with aligned depth using a local robust estimator: a $5{\times}5$ neighborhood around it $(u_c, v_c)$ is sampled from the depth image, and the median of valid non-zero measurements is used as depth~$z$. The camera-frame point is then recovered with the pinhole model~\cite{Szeliski2022CVAA,MurArtal2017ORBSLAM2},
\begin{equation}
  x_c = \frac{(u_c - p_x)\,z}{f_x}, \quad
  y_c = \frac{(v_c - p_y)\,z}{f_y}, \quad
  z_c = z,
\end{equation}
where $f_x, f_y$ are focal lengths and $p_x, p_y$ are principal-point parameters obtained online from \texttt{sensor\_msgs/CameraInfo} which comes from the camera node. This yields $\mathbf{p}_c = [x_c,\,y_c,\,z_c]^\top$ in the camera optical frame.

To obtain executable robot coordinates, $\mathbf{p}_c$ is transformed with an offline ArUco-marker calibration and fixed robot offsets,
\begin{equation}
  \mathbf{p}_{aruco} = \mathbf{R}_{a \leftarrow c}\,\mathbf{p}_c + \mathbf{t}_{a \leftarrow c}, \qquad
  \mathbf{p}_{robot} = \boldsymbol{\Phi}(\mathbf{p}_{aruco}) + \mathbf{d},
\end{equation}
where $\mathbf{R}_{a \leftarrow c}$ and $\mathbf{t}_{a \leftarrow c}$ are loaded from calibration files, $\boldsymbol{\Phi}$ denotes the axis remapping/sign correction to the FR3 convention, and $\mathbf{d}$ is a constant offset vector. The final target is published in the FR3 base frame (\texttt{fr3\_link0}) and mirrored to the web dashboard with pixel and depth, along with the output confidence for explicit user confirmation before motion execution. This step is especially important in overlapped, multi-object scenes and handover tasks where grounding errors could otherwise lead to unsafe or unintended motions.

It is important to note that neither the LLM nor the VLM in this framework has been fine-tuned on manipulation-specific data or task-specific object categories. Both operate exclusively on their general pre-trained knowledge in a zero-shot setting. This is a key distinction from traditional robot perception pipelines that rely on object detectors such as YOLO~\cite{Redmon2016YOLO}. A YOLO-based approach trains a dedicated model on a fixed set of predefined classes, which means it can detect objects within its training distribution efficiently but cannot generalize to new categories without collecting new training data and retraining the model. The VLM used here, by contrast, can ground arbitrary natural-language descriptions, including general objects and relative spatial references, without any retraining. However, in order to reach the highest accuracy in a model, there is always a need for fine-tuning the model.

\subsection{Real-Time Distributed Communication}

All modules communicate through ROS~2 topics~\cite{Macenski2022}, which provide a lightweight and modular messaging interface between perception, reasoning, and the robot. The web dashboard communicates with the ROS~2 backend through a web handler and ROSBridge, enabling real-time status updates, chat interaction, visual feedback, and confirmation dialogs. Motion execution feedback is returned through status topics so that the user interface can display whether the system is waiting, processing, executing, or completed.

The communication design also makes the system easy to test and extend. Individual components can be replaced or moved to different hardware without redesigning the overall pipeline. For example, the language understanding node could be swapped with a different LLM or a fine-tuned model without affecting the vision or motion nodes. Similarly, the vision module could be replaced with a different VLM or a traditional detector if desired. The modular design also supports future extensions such as adding more complex dialogue management, multi-turn interactions, or additional sensing modalities, without requiring a complete change of the framework architecture.

\section{Evaluation}
This section evaluates the proposed framework through a compact benchmark across pick, place, and handover tasks, followed by all representative scene examples. The benchmark checks whether the full pipeline, from natural-language input to robot planning, stays reliable as scene ambiguity increases. 
\subsection{Experimental Workcell}
The experimental platform consists of a Franka Emika FR3 7-degree-of-freedom (7-DoF) robotic arm with a parallel gripper~\cite{Franka2026}, an Intel RealSense D435i RGB-D camera mounted at the end effector~\cite{IntelRealSense2026}, and a web-based dashboard for natural language interaction and motion confirmation. The software stack is built on ROS~2 Jazzy~\cite{Macenski2022} and MoveIt2~\cite{MoveIt2026}, while local model inference is served through Ollama~\cite{Ollama2026}.

\begin{table*}[t]
  \caption{Benchmark summary across pick, place, and handover tasks. Success entries report completed trials over 5 in the format single/multiple/overlapped for each task. \textbf{LLM/VLM lat.} report average per-request latency in seconds from Ollama's \texttt{total\_duration}. \textbf{LLM/VLM file mem.} report quantized weight file size from \texttt{ollama list} (GB). \textbf{LLM/VLM RAM} report observed runtime RAM while serving requests (GB) and observed from \texttt{ollama ps}. VLM execution is restricted to the RTX~6000 workstation for all listed configurations, and \textit{(Jetson)} marks rows in which the LLM is deployed on the NVIDIA Jetson AGX Orin 64 GB platform.}
  \label{tab:comparison}
  \centering
  \setlength{\tabcolsep}{2.8pt}
  \footnotesize
  \resizebox{\textwidth}{!}{%
  \begin{tabular}{|l|l|c|c|c|c|c|c|c|c|c|}
    \hline
    \textbf{LLM model} & \textbf{VLM model} & \textbf{Pick S/M/O} & \textbf{Place S/M/O} & \textbf{Handover S/M/O} & \textbf{LLM lat.} & \textbf{VLM lat.} & \textbf{LLM file} & \textbf{VLM file} & \textbf{LLM RAM} & \textbf{VLM RAM} \\ \hline
    \shortstack[l]{ministral-3:8b\\(Jetson)~\cite{Mistral8B2026}} & \shortstack[l]{Qwen2.5-VL 32B\\~\cite{Qwen25VL2026}} & \textbf{5/5/5} & \textbf{5/5/5} & \textbf{5/5/5} & \textbf{5.47} & \textbf{9.57} & 6.0 & 21.0 & 43.00 & 127.69 \\ \hline
    \shortstack[l]{LLaMA 3.1 8B\\~\cite{Llama31_2026}} & \shortstack[l]{Qwen2.5-VL 32B\\~\cite{Qwen25VL2026}} & 5/5/4 & 5/5/5 & 4/4/4 & 5.97 & 42.83 & 4.9 & 21.0 & 28.58 & 127.69 \\ \hline
    \shortstack[l]{Qwen2.5 7B\\~\cite{Qwen25_2026}} & \shortstack[l]{Llama 3.2-Vision 90B\\~\cite{Llama32Vision2026}} & 1/0/3 & 2/1/2 & 3/2/0 & 5.95 & 40.69 & 4.7 & 54.0 & 7.65 & 116.98 \\ \hline
    \shortstack[l]{DeepSeek-R1 7B\\~\cite{DeepSeekR12026}} & \shortstack[l]{LLaVA 7B\\~\cite{LLaVA2026}} & 0/0/0 & 0/0/0 & 0/0/0 & \texttt{--} & \texttt{--} & 4.7 & 4.7 & 18.33 & 9.85 \\ \hline
\end{tabular}
}
\end{table*}

The default deployed system\footnote{\url{https://github.com/cogrob-tuni/franka-llm}} uses \mbox{ministral-3:8b} on the NVIDIA Jetson AGX Orin 64 GB platform for command understanding~\cite{Mistral8B2026} and Qwen2.5-VL~32B on the RTX~6000 workstation for visual grounding~\cite{Qwen25VL2026}. This pair is used as the default because it gave the best overall balance of accuracy and latency in our experiment. Each scene contains one or a couple of dice with different colors and visible face values, which makes it possible to test language understanding, visual disambiguation, and end-to-end execution under object-level ambiguity.

To study the effect of model choice and hardware placement, the benchmark also includes alternative deployable pairs. In all runs, the VLM stays on the RTX~6000 workstation, while the LLM is deployed either on the RTX~6000 or on the NVIDIA Jetson AGX Orin 64 GB platform.

\subsection{Performance Evaluation Protocol}

We evaluate the system through a repeatability and timing study across pick, place, and handover tasks. The goal is not to rank language or vision models in isolation but to compare complete system configurations that combine an LLM, a VLM, and their deployment hardware. Each configuration is tested with the same scene layouts and the same prompts.

\begin{itemize}
\item \textbf{Scene coverage}: Each task is tested under three scene conditions: single-object, multiple-object, and overlapped-object layouts.
\item \textbf{Trials}: Every task-scene pair is repeated 5 times, which gives 45 runs per model configuration ($3$ tasks $\times$ $3$ scene conditions $\times$ $5$ trials). With four model configurations, the full benchmark includes 180 runs.
\item \textbf{Verbatim prompts}: Pick uses \textit{"pick the yellow dice"}, \textit{"grab the yellow dice"}, \textit{"pick up the yellow dice not the red dice"}, \textit{"can you pick up the yellow dice"}, and \textit{"I need you to pick the yellow dice"}. Place uses \textit{"place it to the left of the red dice"}, \textit{"put it to the right of the red dice"}, \textit{"place it above the red dice"}, \textit{"place it below the red dice"}, and \textit{"put it right side of the red dice"}. Handover uses \textit{"give it to me"}, \textit{"hand it over"}, \textit{"pass it to me"}, \textit{"can you give me the dice"}, and \textit{"deliver the dice carefully to my hand"}.
\item \textbf{Success criterion}: A run is counted as successful only if the correct dice is grounded and the requested pick, place, or handover action is completed.
\item \textbf{Memory metrics}: We report two memory views for both LLM and VLM. (1) \textbf{File-size memory} from \texttt{ollama list}, which is the quantized model weight size on disk (GB). (2) \textbf{Runtime RAM}, measured while Ollama keeps the model loaded in memory, serves requests and refers to the live footprint of models loaded for inference (observed from \texttt{ollama ps}).
\item \textbf{Latency metric}: LLM and VLM latency are reported separately using Ollama's \texttt{total\_duration}. End-to-end latency for one request is therefore $T_{e2e} = T_{LLM} + T_{VLM}$.
\end{itemize}

During evaluation, the dashboard overlays the grounded target and reports pixel, depth, confidence, and robot-frame values before user confirmation and execution. Table~\ref{tab:comparison} summarizes the benchmark configurations and reports success as completed trials out of 5 in the format S/M/O (single/multiple/overlapped), alongside per-module latency and memory for the LLM and VLM.
\section{Results}
\subsection{Experiment Results and Analysis}

The proposed framework is designed for distributed deployment, where reconfiguration is achieved by launching ROS~2 nodes on the designated target devices.
Table~\ref{tab:comparison} summarizes success, latency, and memory for four LLM--VLM configurations (180 total runs) evaluated on pick, place, and handover under single/multiple/overlapped scenes. The default distributed deployment, \mbox{ministral-3:8b} hosted on the Jetson AGX Orin and Qwen2.5-VL~32B hosted on the RTX~6000, achieved full success across all tasks and ambiguity levels (5/5/5 for pick, place, and handover). It also resulted in the lowest reported latencies (5.47~s LLM and 9.57~s VLM), which is due to running each on a different node.

\begin{figure*}[t]
  \centering
  \subcaptionbox{Pick up -- single object\label{fig:pickup_single}}[0.325\textwidth]{%
    \includegraphics[width=\linewidth]{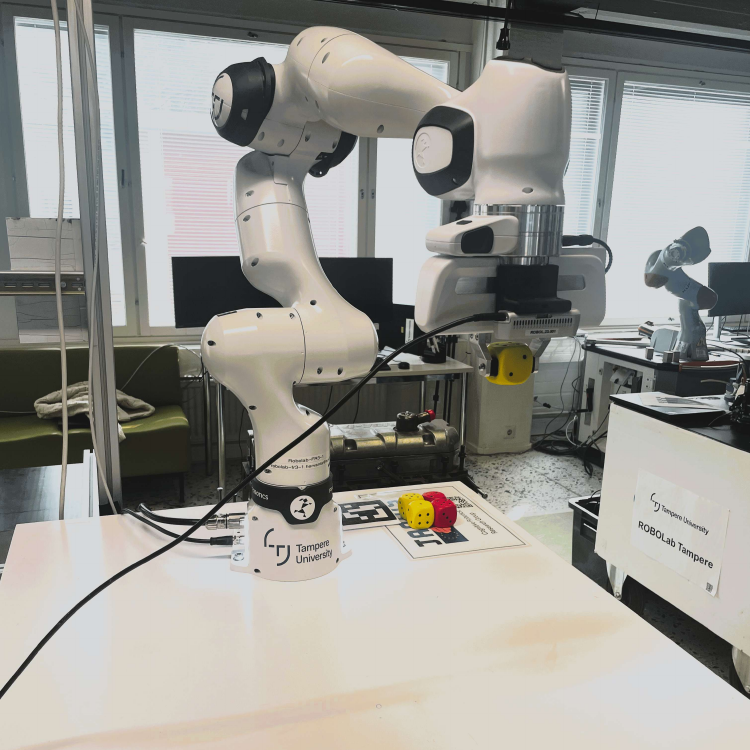}}
  \hfill
  \subcaptionbox{Pick up -- multiple objects\label{fig:pickup_multiple}}[0.325\textwidth]{%
    \includegraphics[width=\linewidth]{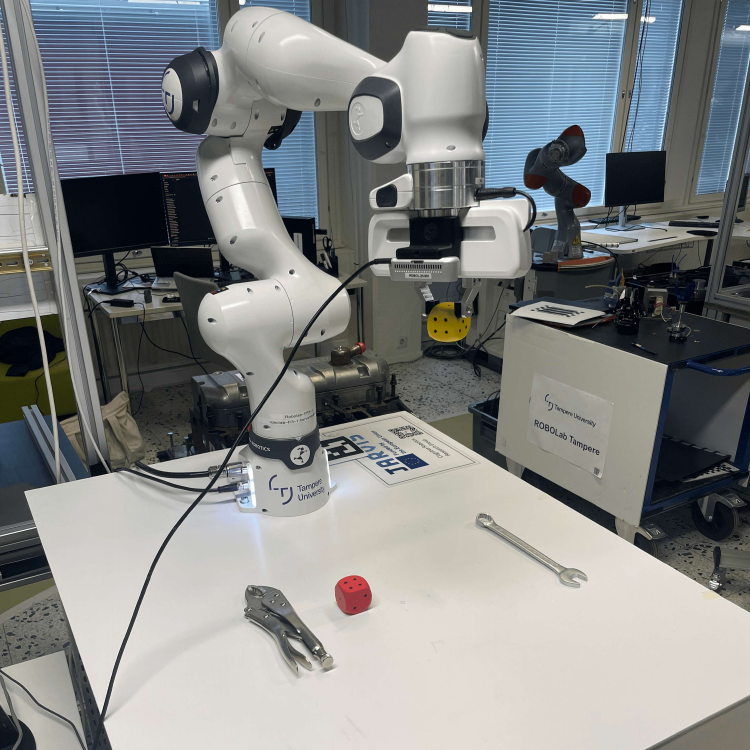}}
  \hfill
  \vspace{6mm}
  \subcaptionbox{Pick up -- overlapped objects\label{fig:pickup_overlapped}}[0.325\textwidth]{%
    \includegraphics[width=\linewidth]{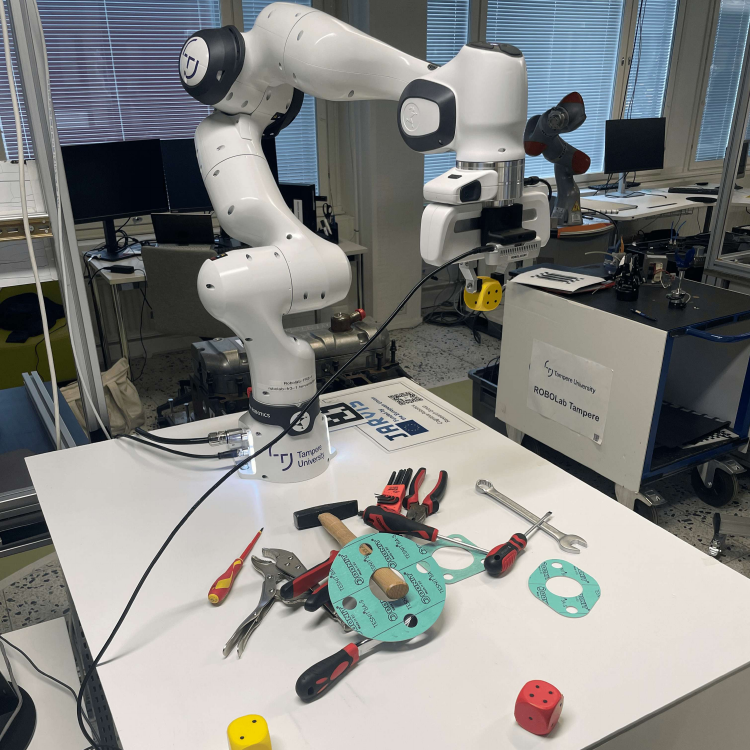}
    } \vskip\floatsep
  \subcaptionbox{Place -- single object\label{fig:place_single}}[0.325\textwidth]{%
    \includegraphics[width=\linewidth]{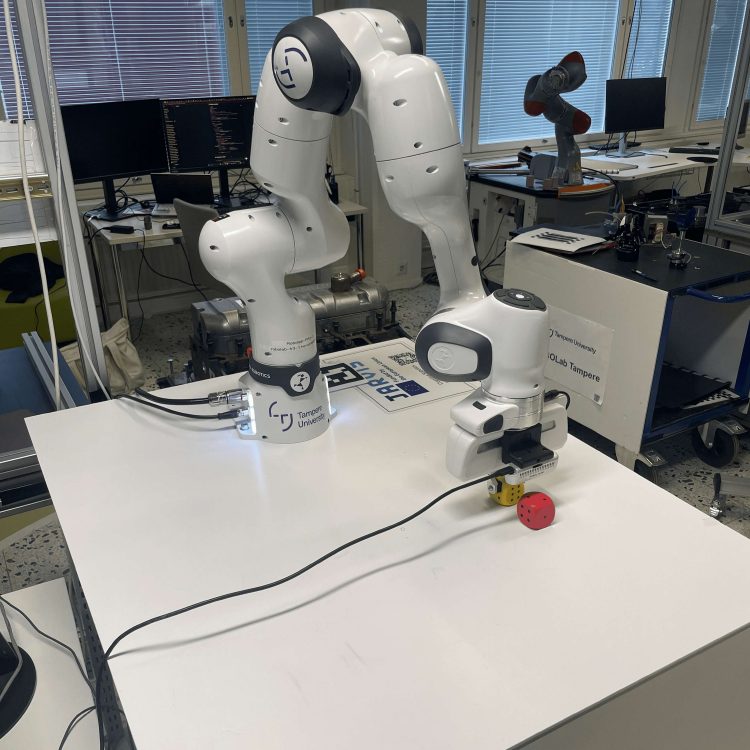}}
  \hfill
  \subcaptionbox{Place -- multiple objects\label{fig:place_multiple}}[0.325\textwidth]{%
    \includegraphics[width=\linewidth]{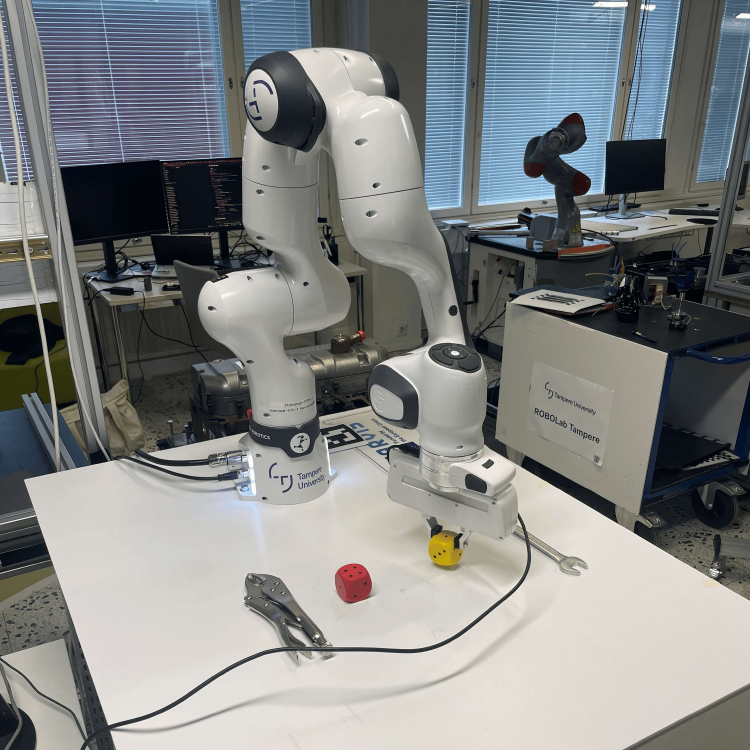}}
  \hfill
  \vspace{8mm}
  \subcaptionbox{Place -- overlapped objects\label{fig:place_overlapped}}[0.325\textwidth]{%
    \includegraphics[width=\linewidth]{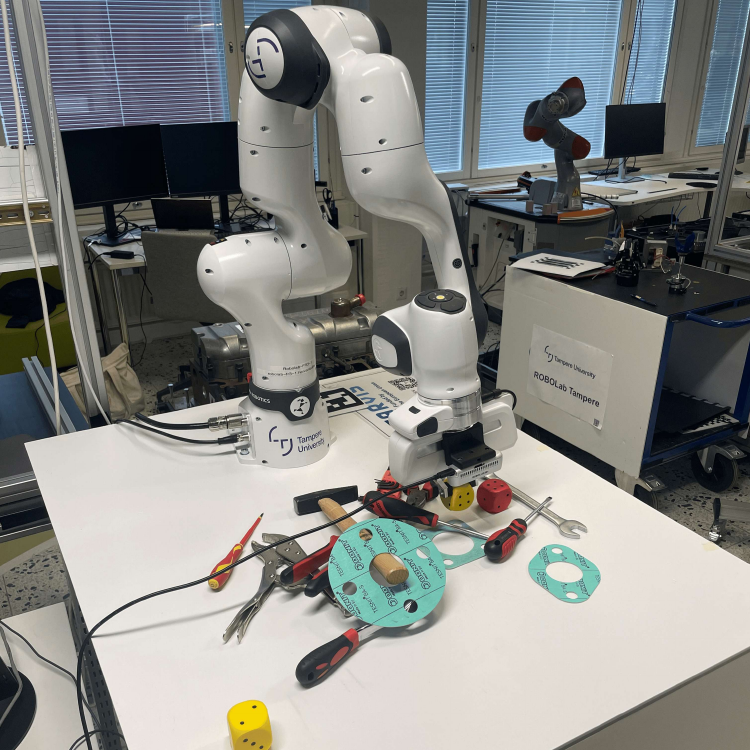}}
    \vskip\floatsep
  \subcaptionbox{Handover -- Single Object\label{fig:handover_single}}[0.325\textwidth]{%
    \includegraphics[width=\linewidth]{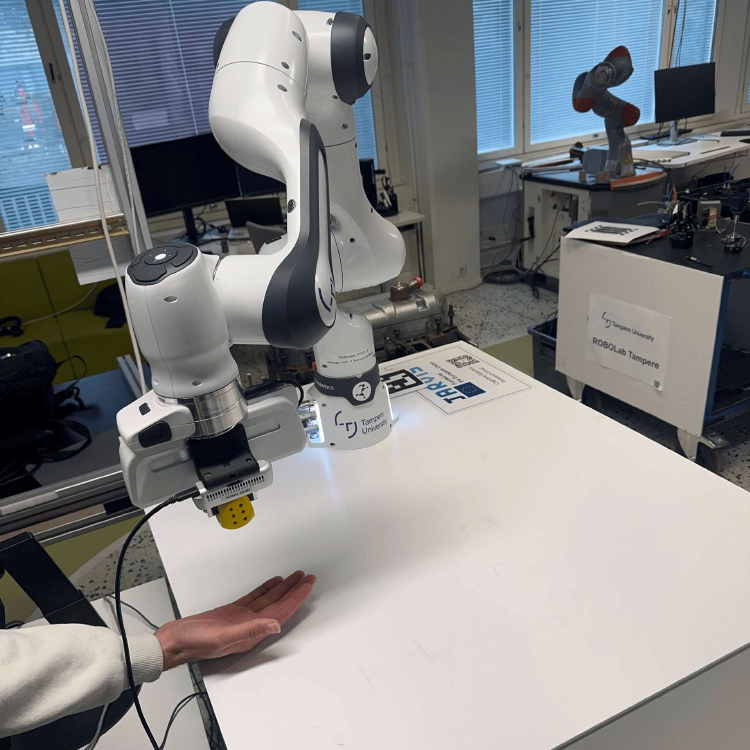}}
  \hfill
  \subcaptionbox{Handover -- Multiple Objects\label{fig:handover_multiple}}[0.325\textwidth]{%
    \includegraphics[width=\linewidth]{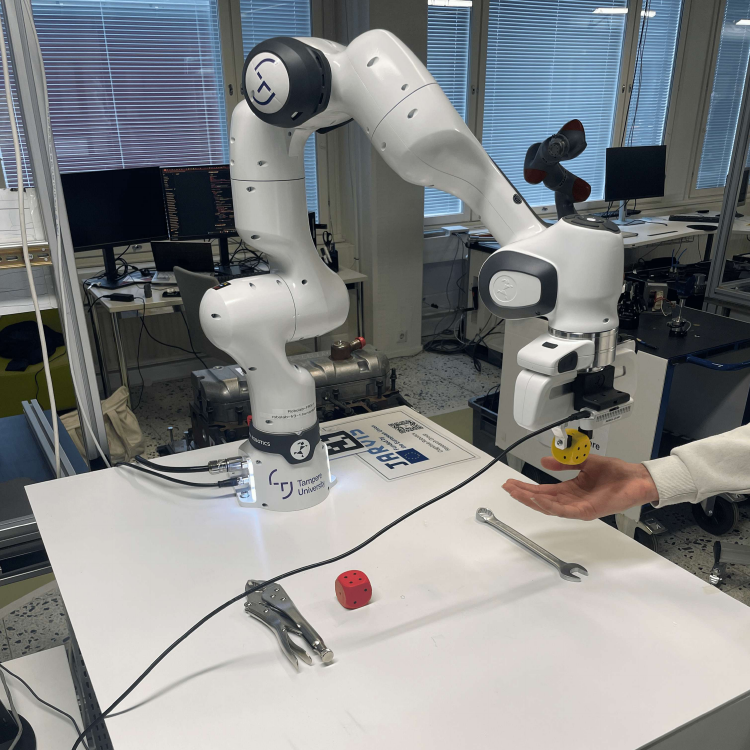}}
  \hfill
  \subcaptionbox{Handover -- Overlapped Objects\label{fig:handover_overlapped}}[0.325\textwidth]{
    \includegraphics[width=\linewidth]{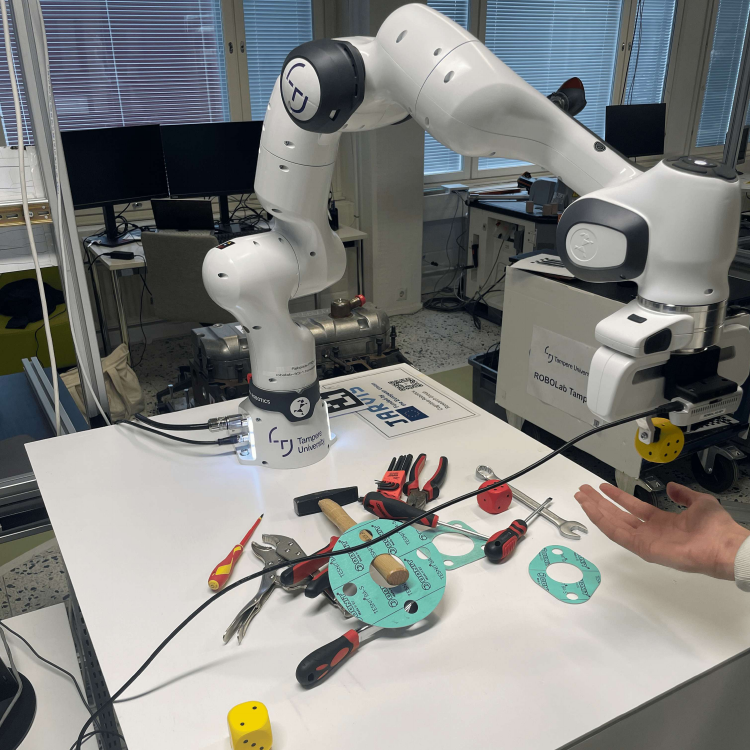}}
\caption{Representative scenes from the experimental evaluation. The top row shows pick examples, the middle row shows place examples, and the bottom row shows handover examples under increasing scene ambiguity. These images correspond to the same benchmark conditions summarized in Table~\ref{tab:comparison}. \label{fig:results}
}
\end{figure*}

\begin{figure*}[t]
  \centering
  \subcaptionbox{Successful grounding example\label{fig:featured_reasoning}}[0.48\textwidth]{%
    \includegraphics[width=\linewidth]{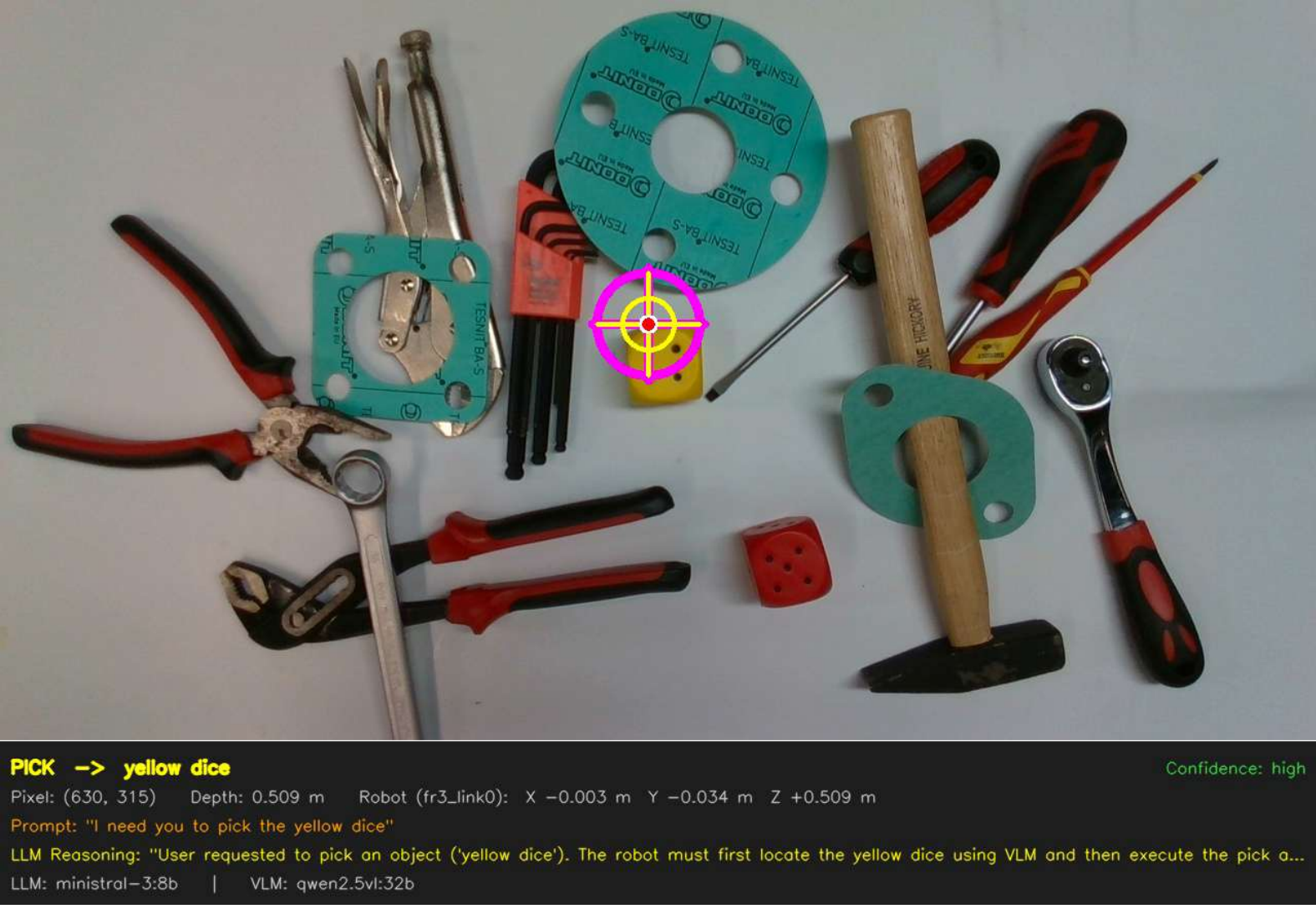}}
  \hfill
  \subcaptionbox{Failure grounding example\label{fig:featured_failure}}[0.48\textwidth]{%
    \includegraphics[width=\linewidth]{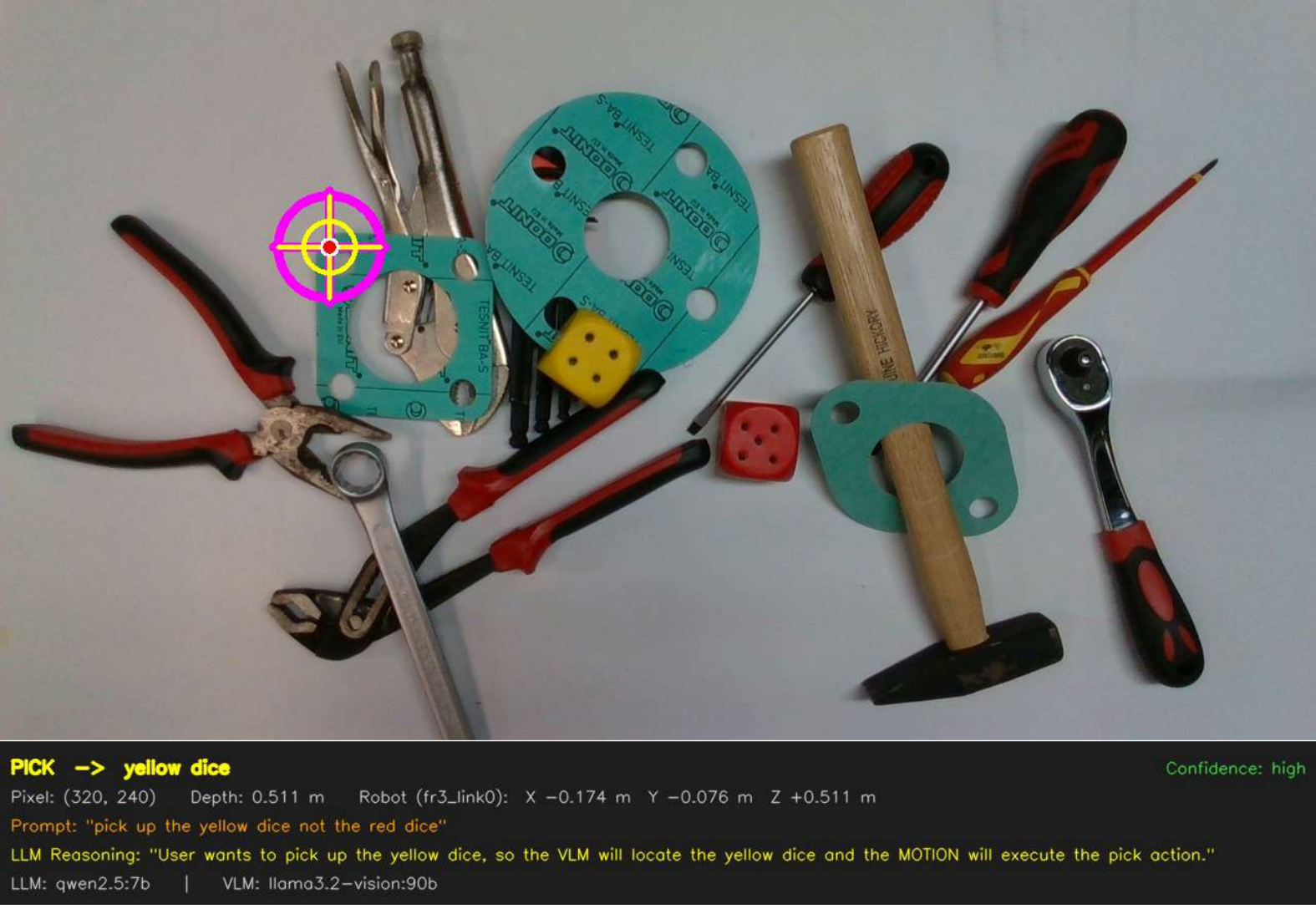}}
  \caption{Grounding outcomes from the deployed setup in an overlapped scene. (a) shows a representative correct grounding result. (b) shows a failure case where the VLM misidentified the target, which motivates keeping explicit user confirmation before any motion is executed.}
  \label{fig:grounding_examples}
\end{figure*}

In contrast, changing the model pairing or where inference runs degrades either robustness or responsiveness. LLaMA~3.1~8B with the same Qwen2.5-VL~32B VLM remained strong on pick and place but dropped on handover and overlapped scenes (4/4/4 on handover), and its VLM latency increased substantially (42.83~s), consistent with higher contention when multiple heavy components share the same workstation. The configuration with a larger VLM (Llama~3.2-Vision~90B) did not improve success: paired with Qwen2.5~7B, it produced inconsistent grounding requests and low success rates across tasks. Finally, DeepSeek-R1~7B with LLaVA~7B failed to complete any trials (0/0/0 across tasks), with many runs not reaching execution.

The memory measurements in Table~\ref{tab:comparison} explain why distributed deployment is practical and often necessary: runtime RAM is dominated by the VLM (e.g., Qwen2.5-VL~32B at 127.69~GB), which motivates hosting vision grounding on a GPU-based machine. By comparison, the LLM planner footprint is smaller and, in the default deployment, fits within the 64~GB memory budget of the Jetson (43.00~GB observed while serving requests), supporting the design choice of pushing language parsing to the edge while reserving workstation resources for vision.

Crucially, the pipeline itself remains unchanged across these configurations: the same ROS~2 modules produce the same typed action primitives and intermediate targets (pixel, depth, and robot-frame) for operator inspection. Model choice and placement are explicit deployment decisions (where nodes are launched) rather than automatic scaling mechanisms.

Fig.~\ref{fig:grounding_examples} provides qualitative context for these quantitative outcomes. Fig.~\ref{fig:grounding_examples}(a) shows a representative successful run in which the grounded target matches the requested object, and the system proceeds after confirmation, while Fig.~\ref{fig:grounding_examples}(b) illustrates a failure mode where the VLM grounds the wrong object despite a high-confidence response. This supports the need for exposing intent/grounding overlays and retaining an explicit confirmation gate before any motion is executed.

\subsection{Limitations}

The system performs effectively for HRI commands involving picking, placing, and handing over objects. However, it can fail on more complex phrasing, for example, commands that require compositional reasoning, such as \textit{"Pick up the dice only if its value is higher than the one next to it"}. Moreover, extremely informal language or combining several instructions into one command can also disrupt the action extraction process, especially when the user requests two actions simultaneously. Additionally, overall accuracy is tied to the capabilities of the deployed models, so there is an inherent trade-off between reliability and inference speed. The system currently does not maintain a conversation history between interactions. Each command is processed independently, i.e., the LLM and VLM have no awareness of the user's previous requests within the same session. This limits the ability to handle follow-up instructions or context-dependent commands such as \textit{"do it again but with the red one instead."} Adding a structured conversational mechanism is a straightforward improvement that could address this.

\subsection{Future Work}

Because generative AI models are trained on broad, general-purpose datasets, a logical next step is to extend them beyond the specific object categories assessed in this benchmark. Through domain-specific fine-tuning of the generative models, especially VLMs, the system could be adapted to specialized use cases beyond picking a dice, such as identifying and manipulating components in assembly workflows (\textit{e.g.,} \textit{"pass me the smaller torque wrench"}), sorting items in logistics environments, or assisting in laboratory settings with custom labware. Fine-tuning the LLM on task-specific manipulation commands would also improve handling of ambiguous inputs and domain terminology. Extending the confirmation step into a clarification dialogue, where the system proactively asks the user a question when intent is uncertain, would further improve robustness without sacrificing safety.

\section{CONCLUSIONS}
We proposed a modular conversational framework that
utilizes local language understanding and vision-language
grounding in a ROS 2 multi-agent pipeline for human-robot
collaborative manipulation tasks. It decouples
high-level reasoning, vision grounding, coordination, and
motion execution so that each module may be executed on
edge platforms based on availability. Our framework also includes a coordinator that translates free-form user commands to structured action
requests, provides pixel/depth/robot coordinates to the
user, and requires explicit user confirmation before executing
motion commands. Our compact benchmark on a Franka
FR3 robot with a Jetson-RTX deployment running
minstral-3:8b and Qwen2.5-VL 32B yields a good tradeoff
between latency and vision grounding for pick, place, and
handover tasks. Our framework provides a practical route
for developing safer and more interpretable language-based
manipulation for HRI in the real world.
\section*{ACKNOWLEDGMENT}
This work was supported by the Research Council of Finland under Grant no. 369003 (PERFORM).

Language refinement was assisted by \textit{Claude Sonnet 4.5}, and Figure~\ref{fig:intro_img} was improved using \textit{GPT Image 1.5}.
\bibliographystyle{IEEEtran}
\nocite{*}
\bibliography{IEEEabrv,refs}

\end{document}